\documentclass[letterpaper, 10 pt, conference]{ieeeconf}
\IEEEoverridecommandlockouts
\usepackage{adjustbox}
\usepackage{graphics}
\usepackage{times}
\usepackage{amsmath}
\usepackage{amssymb}
\usepackage{graphicx}
\usepackage{bm}
\usepackage{algorithm}
\usepackage{siunitx}
\usepackage[noend]{algpseudocode}
\usepackage{balance}
\usepackage{cuted}
\usepackage{capt-of}
\usepackage{booktabs}
\usepackage{xcolor}
\usepackage{standalone}
\usepackage{tikz}
\usetikzlibrary{arrows.meta, positioning, shapes.geometric, calc, fit, backgrounds, shadows}
\graphicspath{{figures/}}
\usepackage{standalone}
\usepackage{pgfplots}
\usepackage{mathtools}
\usepackage{booktabs}
\usepackage{tabularx}
\usepackage{multirow}
\usepackage{makecell}
\usepackage{cite}

\providecolor{BackboneColor}{RGB}{242, 245, 249}
\providecolor{EncoderColor}{RGB}{52, 73, 94}
\providecolor{DecoderColor}{RGB}{26, 188, 156}
\providecolor{LossColor}{RGB}{192, 57, 43}
\providecolor{FeatureColor}{RGB}{241, 196, 15}
\providecolor{OcclColor}{RGB}{142, 68, 173}
\usepackage{textcase}
\usepackage[font=small]{caption}
\usepackage{url}

\def\secref#1{Sec.~\ref{#1}}
\def\figref#1{Fig.~\ref{#1}}
\def\tabref#1{Tab.~\ref{#1}}
\def\eqref#1{Eq.~(\ref{#1})}

\newcommand\etal{\emph{et al.}}

\title{\LARGE \bf
SG-DOR: Learning Scene Graphs with Direction-Conditioned\\Occlusion Reasoning for Pepper Plants}
\author{Rohit Menon$^{*}$ \and Niklas Mueller-Goldingen$^{*}$ \and Sicong Pan  \and Gokul Krishna Chenchani \and Maren Bennewitz
}
\begin{document}
\maketitle
 \def\thefootnote{}\footnotetext{All authors are with the Humanoid Robots Lab, University of Bonn, Germany.
 R. Menon and M. Bennewitz are further affiliated with the Lamarr Institute and the Center for Robotics, Bonn.\\
 This work has partially been funded by the Deutsche Forschungsgemeinschaft (DFG, German Research Foundation) under grant 459376902 – AID4Crops, under Germany’s Excellence Strategy, EXC-2070 -- 390732324 -- PhenoRob, and the German Federal Ministry of Research, Technology and Space~(BMFTR) under the Robotics Institute Germany (RIG).\\
 $^{*}$ denotes equal contribution
 }

\renewcommand{\thefootnote}{\arabic{footnote}}
\begin{abstract}
Robotic harvesting in dense crop canopies requires effective interventions that depend not only on geometry, but also on explicit, direction-conditioned relations identifying which organs obstruct a target fruit.
We present SG-DOR (Scene Graphs with Direction-Conditioned Occlusion Reasoning), a relational framework that, given instance-segmented organ point clouds, infers a scene graph encoding physical attachments and direction-conditioned occlusion.
We introduce an \emph{occlusion ranking} task for retrieving and ranking candidate leaves for a target fruit and approach direction, and propose a direction-aware graph neural architecture with per-fruit leaf-set attention and union-level aggregation.
Experiments on a multi-plant synthetic pepper dataset show improved occlusion prediction (F1=0.73, NDCG@3=0.85) and attachment inference (edge F1=0.83) over strong ablations, yielding a structured relational signal for downstream intervention planning.
\end{abstract}
\section{Introduction}
In precision horticulture, harvesting is typically preceded by monitoring and targeted interventions, such as assessing fruit visibility and inspecting peduncle attachments.
Dense foliage induces strong self-occlusions: fruits are partially hidden by leaves, peduncles are not fully observable, and feasible grasp trajectories are obstructed by surrounding plant organs.
Sweet pepper plants exemplify such articulated, highly occluded structures.
In practice, robots often reposition or remove specific leaves to reveal targets and create safe access, making it essential to identify which organs obstruct a fruit rather than interacting blindly.
While state-of-the-art fruit mapping pipelines detect and map plant instances~\cite{wang2017tree}, they operate at the object level without encoding structural attachments or relational dependencies.

Recent approaches mitigate occlusions via active perception~\cite{burusa2024attention, isaacjose2025iros} or amodal completion~\cite{pan2023panoptic, menon2023nbv, li2025semp}.
Compared to static, passive mapping, they improve geometric fidelity and surface coverage under occlusion and can provide instance-segmented organ hypotheses for downstream reasoning and planning.
However, they do not explicitly identify or rank which organs obstruct a target from a given approach direction.
Such occluder-specific reasoning is required for downstream actions such as targeted leaf pushing, safe pruning, and fruit harvesting; however, existing interactive pipelines typically require the relevant occluding leaf to be known a priori~\cite{yao2025safe}.
This creates a representation gap: improved reconstructions still lack explicit, direction-conditioned identification and ranking of target-occluding organs, motivating structured scene representations for actionable manipulation.
\begin{figure}
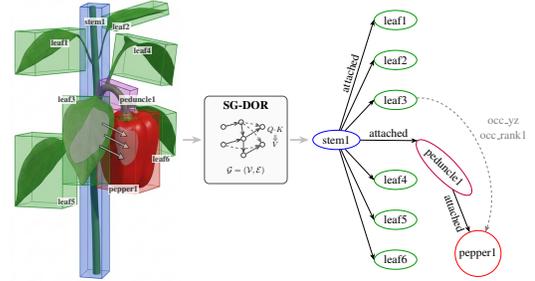

\centering
\includestandalone[width=0.80\linewidth] {figures/full_pipeline_figure}
\caption{Overview of SG-DOR.
Left: Instance-segmented 3D sweet pepper plants.
Center: The SG-DOR network performs attention-based relational reasoning over the induced graph representation.
Right: Inferred scene graph encoding structural attachments and direction-conditioned occlusion ranking between plant organs.
}
\label{fig:full_pipeline}
\vspace{-0.3cm}
\end{figure}

In this work, we propose SG-DOR (\textbf{S}cene \textbf{G}raphs with \textbf{D}irection-Conditioned \textbf{O}cclusion \textbf{R}easoning), a downstream relational reasoning module for pepper plants under severe occlusion.
Assuming a set of instance-segmented 3D plant organs~\cite{dong2025automatic}, provided either by simulation ground truth or an upstream perception pipeline (\figref{fig:full_pipeline}), SG-DOR learns an explicit scene graph, where nodes correspond to individual organs, while edges encode both structural attachments and direction-conditioned 3D occlusion relations.
By defining these relations in an object-centric, fruit-local coordinate frame, occlusion reasoning remains consistent across viewing directions.
This enables future occlusion-aware manipulation strategies by explicitly identifying and ranking candidate occluders. The main contributions of this work are:

\begin{itemize}
\item We formulate direction-conditioned 3D occlusion reasoning as a downstream relational learning problem and introduce an occlusion-aware scene graph that jointly encodes organ attachments and per-direction occluder rankings from instance-segmented organ point sets.
\item We propose SG-DOR, a direction-aware graph neural network with per-fruit leaf-set attention and union-level aggregation for joint attachment inference and rank-aware occlusion prediction.
\item We develop a biologically consistent procedural generation and annotation pipeline and release a large-scale synthetic multi-plant pepper dataset with ground-truth directional occlusion labels for scalable supervision and benchmarking of future perception reasoning systems.
\end{itemize}

Experiments validate that SG-DOR can reliably infer direction-conditioned occlusion relations and occluder rankings while preserving structural attachment prediction.
The code, dataset-generation pipeline, training configuration, and supplementary documentation are publicly available at \url{https://github.com/HumanoidsBonn/SG-DOR}.

\section{Related Work}
Occlusion challenges robotic perception and reasoning, degrading performance in instance segmentation~\cite{qi2022occluded}, motion planning~\cite{wang2023occlusion}, and object tracking~\cite{liang2018tracking}.
While amodal scene analysis has advanced completion-based instance segmentation~\cite{zhang2024amodal}, and depth ordering via occlusion graphs captures pairwise relations~\cite{ming2015monocular}, these methods predict binary occlusion in 2D images.
Unlike SG-DOR, they do not perform direction-conditioned occlusion ranking.

To mitigate occlusion in embodied settings, existing active and interactive perception methods rely on belief models~\cite{eidenberger2009probabilistic} and heuristic visibility corridors~\cite{ dengler2025efficient} to reduce spatial uncertainty.
More recently, VISO-Grasp~\cite{shi2025viso} leverages foundation models to address occlusions by constructing an instance-centric spatial representation for Next-Best-View planning for improved grasping.
SG-DOR, on the other hand, addresses the fundamental prerequisite of structured relational reasoning by learning an explicit scene graph that maps how occlusion shifts based on the approach trajectory.

In the 3D domain, Armeni~\etal~\cite{armeni20193d} model occlusion as a geometric byproduct within a unified 3D scene graph spanning buildings, rooms, and objects, while Zia~\etal~\cite{zia2015towards} infer part-level occlusion for fine-grained 3D object modeling.
Similarly, although Wald~\etal~\cite{wald2022learning} also rely on point cloud instance embeddings, they restrict their focus to learning semantic relations between objects.
In contrast, SG-DOR repurposes scene graphs specifically for graded occlusion reasoning, enabling explicit identification of which surrounding structures obstruct a target within an object-centric 3D frame.

In agricultural manipulation, dense foliage renders occlusion highly detrimental~\cite{lenz2024hortibot}.
While existing amodal completion~\cite{li2025cga} and active perception~\cite{burusa2024attention, menon2023nbv} frameworks improve target-centric geometric completeness through shape inference and viewpoint optimization, they do not explicitly model the occluding plant organs.
Consequently, under severe occlusion where viewpoint changes alone offer limited benefit, these methods struggle because they cannot identify the specific structures that must be manipulated.
Even when employing hierarchical 3D scene graphs for semantic-metric plant mapping and fruit counting~\cite{yang2025hierarchical}, the focus remains strictly on inspection and monitoring rather than explicit occlusion analysis.
Leaf manipulation and pruning have been explored to improve visibility, but they either assume that the occluding leaf is already identified~\cite{yao2025safe} or resort to exploratory brute-force interaction~\cite{giang2024autonomous}.
All these methods lack a structured representation that quantifies which specific leaves contribute most to fruit occlusion.

To our knowledge, SG-DOR is the first scene-graph framework to learn direction-conditioned occlusion ranking among plant organs in a 3D object-centric frame.

\section{Biologically Consistent Synthetic Multi-Plant Pepper Dataset for SG-DOR}
\label{sec:dataset_generation}
Relational supervision of plant topology and direction-conditioned occlusion is difficult to obtain in real greenhouse data due to severe self-occlusion and limited visibility of attachment points.
We adopt a fully synthetic generation pipeline to obtain complete structural and visibility ground truth while retaining biological plausibility, enabling controlled large-scale supervision of relational and occlusion reasoning.

\subsection{Procedural Topology Generation}
We generate scenes in BlenderProc~\cite{denninger2023BlenderProc2} using curated stem, leaf, peduncle, and fruit prototypes adapted from the WUR synthetic capsicum models~\cite{barth2018data}.
Organs are instantiated at predefined attachment sockets, enforcing biologically consistent topology (stem–leaf, stem–peduncle, peduncle–fruit) by construction.
Multiple stems are arranged in short strips to mimic greenhouse rows.
We apply domain randomization over stem count, organ type, scale, orientation, and attachment height.
To maintain geometric plausibility under dense clutter, candidate placements undergo staged broad-phase and narrow-phase collision checks.
Ground-truth attachment edges are directly obtained from the generated graph.

\subsection{Pointcloud-Based Occlusion Labeling}
\begin{figure}[b]
\centering
\includegraphics[width=\columnwidth]{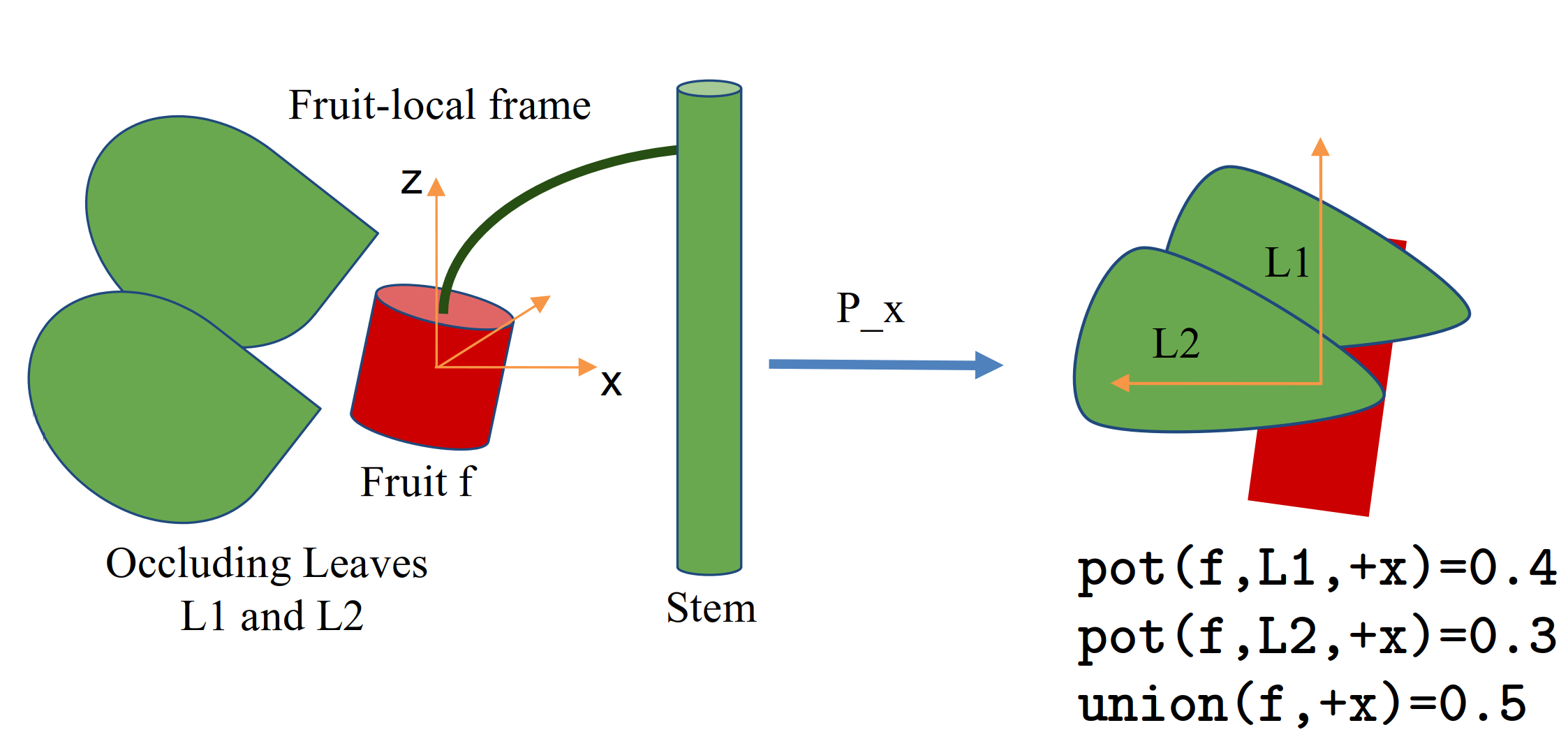}
\caption{
\textbf{Geometric occlusion modeling via depth comparison.}
\scriptsize{L1 and L2 are candidate leaves occluding Fruit f, with L1 closer to the fruit and its potential occlusion score along the +x-axis being 0.4, while that of L2 is 0.3. Due to overlapping of L1 and L2, their union occlusion score along + x-axis is 0.5.}
}
\label{fig:occlusion}
\vspace{-0.2cm}
\end{figure}

\begin{figure*}[t]
\centering
\begin{tikzpicture}[
    node distance=0.5cm,
    font=\small,
    >=Stealth,
    block/.style={
        rectangle, draw=black!80, thick,
        fill=white, rounded corners=2pt,
        minimum width=1.6cm, minimum height=0.6cm,
        align=center, anchor=west, drop shadow={opacity=0.1, shadow xshift=0.3mm, shadow yshift=-0.3mm}
    },
    container/.style={
        rectangle, rounded corners=8pt,
        draw=black!15, fill=#1, inner sep=12pt
    },
    tensor/.style={
        trapezium, trapezium left angle=70, trapezium right angle=110,
        draw=black!80, thick, fill=white,
        minimum width=1.2cm, minimum height=0.6cm, align=center
    },
    enc_node/.style={
        rectangle, draw=black!70, fill=EncoderColor,
        minimum width=0.18cm, rounded corners=0.5pt
    },
    dec_node/.style={
        rectangle, draw=black!70, fill=DecoderColor,
        minimum width=0.18cm, rounded corners=0.5pt
    },
    loss_tag/.style={
        draw=LossColor, fill=LossColor!5, thick, rounded corners=2pt,
        font=\scriptsize\ttfamily, inner sep=2pt, text=LossColor, minimum width=0.85cm
    },
    arrow/.style={thick, ->, black!70},
    skip_arrow/.style={thick, ->, black!50, dashed, rounded corners=5pt}
]

\node (in_pts) [tensor, fill=gray!10, yshift=8.0cm] {$\mathcal{P}$};
\node (in_pts_label) [below=2pt of in_pts, font=\scriptsize\color{black!60}] {$N_p\times 3$};

\node (msg1) [enc_node, right=1.2cm of in_pts, minimum height=1.4cm] {};
\node (msg2) [enc_node, right=0.2cm of msg1, minimum height=1.1cm] {};
\node (msg3) [enc_node, right=0.2cm of msg2, minimum height=0.8cm] {};
\node (msg_label) [below=4pt of msg2, font=\bfseries\scriptsize, align=center] {PointNet++\\MSG};
\node (msg_label_fit) [above=0pt of msg_label, inner sep=0pt] {};

\node (identity) [tensor, right=0.8cm of msg3, fill=FeatureColor!30] {$\mathbf{h}^{\text{id}}$};
\node (local_node_head)   [block, above= 1.9cm of msg3, xshift=1.4cm, minimum width=1.5cm, text width=1.5cm] {\textbf{Semantic}\\$\hat{\bm{y}} \in \mathcal{C}^N$};
\node [loss_tag, right=0.2cm of local_node_head] (l_local_sem) {$\mathcal{L}_{wce}$};
\node (graph_build) [block, below=1.7cm of identity, minimum height=0.8cm] {
    \textbf{Candidate Builder}\\
    \scriptsize $k$-NN + radius + stems
};
\node (edges) [tensor, right=0.8cm of graph_build, fill=FeatureColor!30] {$\bm{\mathcal{E}}$};
\node (edge_label) [below=2pt of edges, font=\scriptsize\color{black!60}] {$E \times 4$};

\node (gine1) [dec_node, right=1.6cm of identity, minimum height=1.4cm, fill=DecoderColor] {};
\node (gine2) [dec_node, right=0.2cm of gine1, minimum height=1.4cm, fill=DecoderColor] {};
\node (gine3) [dec_node, right=0.2cm of gine2, minimum height=1.4cm, fill=DecoderColor] {};
\node (gine_label) [below=4pt of gine2, font=\bfseries\scriptsize, align=center] {ResGINE\\Backbone};
\node (gine_label_fit) [above=0pt of gine_label, inner sep=0pt] {};

\node (h_gnn) [tensor, right=0.6cm of gine3, fill=FeatureColor!30] {$\mathbf{h}^{\text{gnn}}$};
\node (h_gnn_fit) [right=0pt of h_gnn, inner sep=0pt] {};

\node (concat_res) [block, right=0.8cm of h_gnn, fill=gray!5] {
    \textbf{Identity Concat}\\
    $\mathbf{h} = [\mathbf{h}^\mathit{id}, \mathbf{h}^\mathit{gnn}]$
};

\node (node_head)   [block, right=1.5cm of concat_res, yshift=0.0cm, minimum width=3.6cm, text width=3.6cm] {\textbf{Semantic} $\hat{\bm{y}} \in \mathcal{C}^N$};
\node (offset_head) [block, below=0.15cm of node_head, yshift=0.0cm, minimum width=3.6cm, text width=3.6cm] {\textbf{Offset} $\bm{\Delta c} \in \mathbb{R}^3$};
\node (extent_head) [block, below=0.15cm of offset_head, yshift=0.0cm, minimum width=3.6cm, text width=3.6cm] {\textbf{Extent} $\bm{\Delta e} \in \mathbb{R}^3$};
\node (edge_exist)  [block, below=0.15cm of extent_head, yshift=0.0cm, minimum width=3.6cm, text width=3.6cm] {\textbf{Edge Exist} $\hat{e}_{ex} \in [0,1]^E$};
\node (edge_label_head) [block, below=0.15cm of edge_exist, minimum width=3.6cm, text width=3.6cm] {\textbf{Edge Class} $\hat{y}_{ed} \in \mathcal{C}_{e}^E$};
\node (edge_label_head_fit) [below=0pt of edge_label_head, inner sep=0pt] {};

\node [loss_tag, right=0.2cm of node_head] (l_sem) {$\mathcal{L}_{wce}$};
\node [loss_tag, right=0.2cm of offset_head] (l_off) {$\mathcal{L}_{sl1}$};
\node [loss_tag, right=0.2cm of extent_head] (l_ext) {$\mathcal{L}_{sl1}$};
\node [loss_tag, right=0.2cm of edge_exist] (l_edgex) {$\mathcal{L}_{bce}$};
\node [loss_tag, right=0.2cm of edge_label_head] (l_edgel) {$\mathcal{L}_{ce}$};

\node (occl_sa) [block, anchor=west, at={(node_head.west)}, yshift=3.5cm, fill=OcclColor!10, draw=OcclColor, minimum width=2.2cm] {
    \textbf{Leaf SA}\\
    \scriptsize Transformer
};
\node (occl_cross) [block, below=0.4cm of occl_sa, fill=OcclColor!10, draw=OcclColor, minimum width=2.2cm] {
    \textbf{Cross-Attn}\\
    \scriptsize Q: ($\mathbf{h}^\mathit{fruit}, \bm{d}$)\\
    \scriptsize K/V: Leaves
};

\node (pair_pot)  [block, right=1.0cm of occl_sa, yshift=-0.7cm, minimum width=2.1cm] {\textbf{Pair Pot.}\\$\hat{s} \in \mathbb{R}^{M \times K}$};
\node (list_rank)  [block, above=0.5cm of pair_pot, yshift=-0.4cm, minimum width=2.1cm] {\textbf{Pair Rank}\\$\hat{r} \in \mathbb{R}^{M \times K}$};
\node (union_vis) [block, below=0.1cm of pair_pot, minimum width=1.8cm] {\textbf{Union Vis.}\\$\hat{u} \in [0,1]^{I \times K}$};

\node [loss_tag, right=0.2cm of list_rank,  yshift=0.0cm] (l_rank) {$\mathcal{L}_{rank}$};
\node [loss_tag, right=0.2cm of pair_pot, yshift=-0.0cm] (l_pot) {$\mathcal{L}_{bce}$};
\node [loss_tag, right=0.2cm of union_vis, yshift=0.25cm] (l_vis) {$\mathcal{L}_{bce}$};
\node [loss_tag, right=0.2cm of union_vis, yshift=-0.25cm] (l_cons) {$\mathcal{L}_{mse}$};

\draw [arrow] (in_pts) -- (msg1);
\draw [arrow] (msg3) -- (identity);
\draw[arrow] (msg3.east) -| ([xshift=-0.29cm]local_node_head.west) -- (local_node_head.west);;
\draw[arrow] (local_node_head.south) -- (identity.north) ;
\draw [arrow] (identity) -- (graph_build.north);
\draw [arrow] (graph_build) -- (edges);

\draw [arrow] (identity.east) -- (gine1.west);

\draw [arrow] (edges.west) -| ([xshift=-0.29cm, yshift=-0.4cm]gine1.west) -- ([yshift=-0.4cm]gine1.west);

\draw [arrow] (gine3) -- (h_gnn);
\draw [arrow] (h_gnn) -- (concat_res);

\draw [skip_arrow] ([xshift=0.30cm]identity.east) .. controls +(0,3.5) and +(0,2.5) .. (concat_res.north) node[midway, above, font=\scriptsize, align=center] {Identity Skip};

\draw [arrow] (concat_res.east) -- ++(0.3,0) |- (node_head.west);
\draw [arrow] (concat_res.east) -- ++(0.3,0) |- (offset_head.west);
\draw [arrow] (concat_res.east) -- ++(0.3,0) |- (extent_head.west);

\draw [arrow] (h_gnn.south) |- ([yshift=0.2cm]edge_exist.west);
\draw [arrow] (h_gnn.south) |- ([yshift=0.2cm]edge_label_head.west);
\draw [skip_arrow] (edges.east) -| ([xshift=-0.8cm]edge_exist.west) |- ([yshift=-0.2cm]edge_exist.west);
\draw [skip_arrow] (edges.east) -| ([xshift=-0.8cm]edge_label_head.west) |- ([yshift=-0.2cm]edge_label_head.west);

\draw [arrow] (concat_res.east) -- ++(0.3,0) |- ([yshift=0.2cm]occl_sa.west) node[near end, above, font=\scriptsize] {$\mathbf{h}$};
\draw [arrow] (occl_sa) -- (occl_cross);
\draw [skip_arrow] (edges.east) -| ([xshift=-0.8cm, yshift=-0.2cm]occl_sa.west) -- ([yshift=-0.2cm]occl_sa.west) node[midway, below, font=\scriptsize] {Pairs};

\draw [arrow] (occl_cross.east) -- ++(0.3,0) |- (pair_pot.west);
\draw [arrow] (occl_cross.east) -- ++(0.3,0) |- (list_rank.west);
\draw [arrow] (occl_cross.east) -- ++(0.3,0) |- (union_vis.west);

\begin{pgfonlayer}{background}
    \node [container=BackboneColor, fit=(in_pts) (msg_label_fit) (identity), inner xsep=12pt, inner ysep=24pt, label={[anchor=north, font=\bfseries\color{black!70}]north:Organ-Level Encoder}] {};

    \node [container=BackboneColor, fit=(gine1) (gine_label_fit) (h_gnn_fit), inner xsep=16pt, inner ysep=24pt,  label={[anchor=north, font=\bfseries\color{black!70}]north:Relational Backbone}] {};

    \node [rectangle, rounded corners=8pt, fill=LossColor!5, draw=LossColor!20, fit=(node_head) (l_edgel), inner xsep=12pt, inner ysep=13pt, label={[anchor=north, font=\bfseries\color{LossColor}]north:Joint Multi-Task Heads}] {};

    \node [rectangle, rounded corners=8pt, fill=OcclColor!5, draw=OcclColor!20, fit=(occl_sa) (l_vis) (list_rank) (union_vis), inner xsep=12pt,  inner ysep=13pt, label={[anchor=north, font=\bfseries\color{OcclColor}]north:Occlusion Module}] {};
\end{pgfonlayer}
\path[use as bounding box]
  (current bounding box.south west)
  rectangle ([yshift=10pt]current bounding box.north east);
\end{tikzpicture}
\caption{
\textbf{SG-DOR Learning Architecture.}
\scriptsize{Instance-level point sets are encoded by a PointNet++ module to obtain identity embeddings, from which a candidate graph is constructed and refined using a residual GINE backbone with edge features.
Identity and relational features are concatenated and fed to joint multi-task heads for semantic, geometric, and structural predictions.
An additional attention-based occlusion module, shown in detail in \figref{fig:occl_attn}, performs direction-conditioned leaf–fruit reasoning and is trained with ranking and visibility losses. }}
\vspace{-0.4cm}
\label{fig:sg_dor}
\end{figure*}
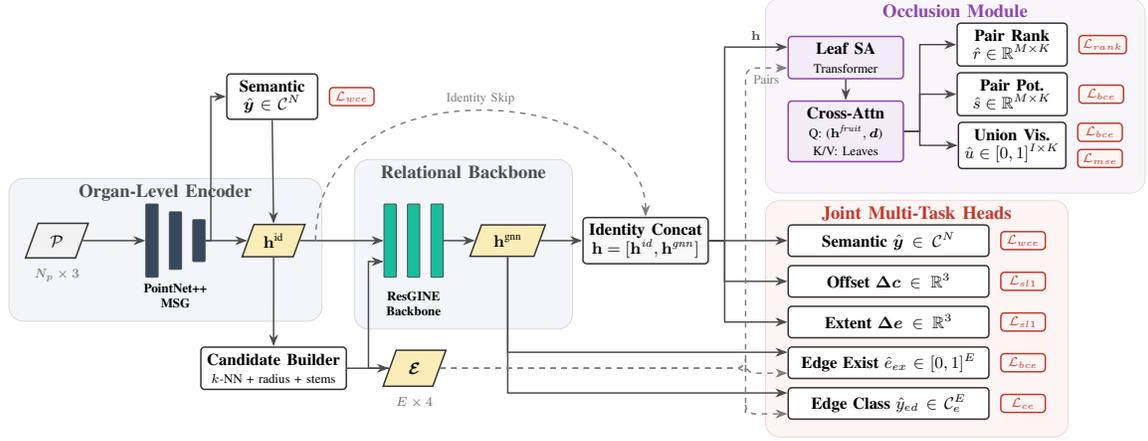
Occlusion is computed from voxelized point clouds of rendered meshes (Fig.~\ref{fig:occlusion}).
We define a fruit-local frame with $\hat{\mathbf{z}}$ as the global up axis, $\hat{\mathbf{x}}$ as the normalized $xy$-projection toward the stem, and $\hat{\mathbf{y}} = \hat{\mathbf{z}} \times \hat{\mathbf{x}}$.
For each fruit, visibility is evaluated along $K=18$ canonical directions: 6 axis-aligned and 12 bi-diagonal directions.
To handle overlapping foliage we compute a depth-ordered z-buffer with $Z=3$ layers. This yields the graded occlusion mass $m_{j,k}$, representing the fraction of fruit voxels for which leaf $j$ appears among the first $Z$ occluding surfaces along direction $\mathbf{d}_k$.
Supervision comprises a fruit-level \emph{union} target (total visibility reduction) and three leaf-level targets: (i) \emph{potential} (any occlusion along $\mathbf{k}$), (ii) \emph{exclusive} (single leaf causing maximal reduction), and (iii) \emph{rank-based} occlusion (graded relevance).
To evaluate the occlusion fraction captured by the top-$Z$ predicted leaves across all directions $K$, we track:
\begin{equation}
\label{eq:zbuf}
\vspace{-0.1cm}
    \text{Mass@}K = \frac{1}{|K|} \sum_{k \in K} \frac{\sum_{j \in \text{top}_{Z}(k)} m_{j,k}}{\sum_{j} m_{j,k} + \epsilon}
\end{equation}

\section{Learning Architecture for SG-DOR}
\label{sec:method}
Reasoning about occlusion in cluttered plant scenes requires a representation that supports both organ-level geometric fidelity and structured relational inference.
We assume instance-segmented organ point clouds are given, and focus on downstream relational reasoning.
We formulate this as multi-task learning~(\figref{fig:sg_dor}) on a directed instance graph $\mathcal{G}=(\mathcal{V},\mathcal{E})$, where nodes correspond to segmented plant organs and edges represent candidate structural and occlusion relations.
The objective is to jointly infer semantic identity, geometric attributes, structural relations, and direction-conditioned occlusion ranking.
\vspace{-0.1cm}

\subsection{Instance-Level Representation}
We represent each node $i \in \mathcal{V}$ using only point coordinates excluding color and surface normals, so that relational inference is driven purely by spatial structure.
Each instance is modeled as an unordered point set $P_i \in \mathbb{R}^{N_p \times 3}$.
We remove shoot-level point clusters from supervision.
These thin structures are often near the sensor resolution limit and difficult to detect reliably.
By omitting them, the model is encouraged to infer structural dependencies directly through stem-leaf relations, promoting more stable relational learning under partial observations.
\vspace{-0.1cm}

\subsection{Local Geometric Encoding}
Each instance $i \in \mathcal{V}$ is represented by a point set $\mathcal{P}_i=\{\mathbf{x}_p\}_{p=1}^{N_p}$ with $\mathbf{x}_p \in \mathbb{R}^3$.
Points are centered per instance by subtracting their mean to encourage translation invariance.
A PointNet++~\cite{qi2017pointnet++} encoder with multi-scale grouping maps the centered set to a permutation-invariant intrinsic geometry embedding $\mathbf{h}^{\text{id}}_i \in \mathbb{R}^{D}$.
This embedding captures the local geometric identity of the organ prior to relational message passing.

\subsection{Local Semantic Prediction Head}
\label{sec:aux_local_node_head}
To guide candidate edge construction without ground-truth labels, we predict the semantic node type~$\mathbf{z}^{\text{local}}_i$ from the intrinsic instance embedding $\mathbf{h}^{\text{id}}_i$ using a
lightweight MLP
\begin{align}
\mathbf{z}^{\text{local}}_i
=
\mathrm{MLP}_{\text{local}}(\mathbf{h}^{\text{id}}_i)
\in \mathbb{R}^{C},
\end{align}
with $C=4$ organ classes (stem, leaf, peduncle, fruit).
The predicted logits enable type-aware edge rules (e.g., stem-specific connections) and act as auxiliary supervision that regularizes the encoder.
They are concatenated with $\mathbf{h}^{\text{id}}_i$ as input to the relational backbone.

\subsection{Candidate Graph Construction}
\label{sec:candidate_graph}
Given the set of nodes $\mathcal{V}$, we construct an over-complete directed candidate edge set $\mathcal{E}_{\text{cand}} = \mathcal{E}_{\text{kNN}} \cup \mathcal{E}_{\text{rad}} \cup \mathcal{E}_{\text{stem}} $,
where proximity-based edges are first proposed using \mbox{$k$-nearest} neighbors ($\mathcal{E}_{\text{kNN}}$) and radius-based search ($\mathcal{E}_{\text{rad}}$) on instance centroids.
These local edges provide a strong geometric prior for structural attachments and line-of-sight occlusions.
However, because stems are vertically elongated, their centroids often reside far from true attachment points, making standard proximity measures poor proxies.
We therefore introduce $\mathcal{E}_{\text{stem}}$, which connects each peduncle and leaf instance to its two nearest stems predicted by the auxiliary head in the global $xy$-plane.

Each candidate edge $(i,j)\in\mathcal{E}_{\text{cand}}$ is augmented with geometric attributes
\begin{equation}
\Delta \mathbf{c}_{ij} = \mathbf{c}_j - \mathbf{c}_i, \quad
d_{ij} = \|\Delta \mathbf{c}_{ij}\|_2
\end{equation}
which encode relative displacement and distance.
These attributes provide the necessary inductive bias for direction-dependent and proximity-dependent relational reasoning.

\subsection{Residual Edge-Aware GNN Backbone}
\label{sec:resgine_backbone}
We refine node embeddings using a residual GINE~\cite{hu2020strategies} based message passing network:
To aggregate information from the local neighborhood $\mathcal{N}_i = \{j \mid (j,i) \in \mathcal{E}_{\text{cand}}\}$, the network performs edge-conditioned updates:
\begin{align}
\mathbf{h}^{\text{gnn}}_i
=
\mathrm{ResGINE}_\theta
\!\left(
\mathbf{h}^{\text{id}}_i, \{ \mathbf{h}^{\text{id}}_j \}_{j \in \mathcal{N}_i}, \mathcal{E}_{\text{cand}}
\right)
\in \mathbb{R}^{G}
\end{align}
In this formulation, $\theta$ represents the learnable weights of the MLP-based message and update functions and $\mathbb{R}^{G}$ denotes the feature dimension of $\mathbf{h}^{\text{gnn}}$.
Edge-conditioned aggregation enables relational reasoning that depends jointly on organ identity and the relative geometry encoded in the candidate edges.
As shown in \figref{fig:sg_dor}, residual connections are employed to mitigate feature over-smoothing in dense foliage and preserve instance-specific geometric information.
The final node representation~$\mathbf{h}_i$ is obtained by concatenating the intrinsic and refined embeddings, utilized by the downstream structural and occlusion heads for task-specific inference:
\begin{align}
\mathbf{h}_i = \big[ \mathbf{h}^{\text{id}}_i, \mathbf{h}^{\text{gnn}}_i \big]
\end{align}

\subsection{Prediction Heads}
As shown in~\figref{fig:sg_dor}, SG-DOR uses multiple prediction heads for node semantics, edge existence and type, geometric regression, and occlusion reasoning (union, pairwise potential, and rank).

\subsubsection{Node Semantics}
A semantic classification head operates on $\mathbf{h}_i$, complementing the auxiliary local classifier by incorporating relational context to disambiguate when geometry alone is insufficient.

\subsubsection{Edge Existence and Relation Type}
For each candidate edge $(u,v)$, we construct a joint edge feature:
\begin{equation}
\mathbf{z}_{u,v} = [\mathbf{h}_u \parallel \mathbf{h}_v \parallel \mathbf{w}_{u,v}]
\end{equation}
with geometric priors $\mathbf{w}_{u,v}$. Parallel heads predict (i) binary edge existence and (ii) multi-class relation type. Separating existence from labeling decouples graph sparsification from semantic identification, reducing gradient interference.

\subsubsection{Geometric Regression}
We regress the centroid offset $\Delta \hat{\mathbf{c}}_i \in \mathbb{R}^{3}$ and 3D extents $\hat{\mathbf{s}}_i \in \mathbb{R}^{3}$ from the node embedding~$\mathbf{h}_i$.
Offsets are predicted relative to the instance cluster mean to improve translation invariance and compensate for biased partial observations.

\subsubsection{Direction-Conditioned Occlusion Reasoning}
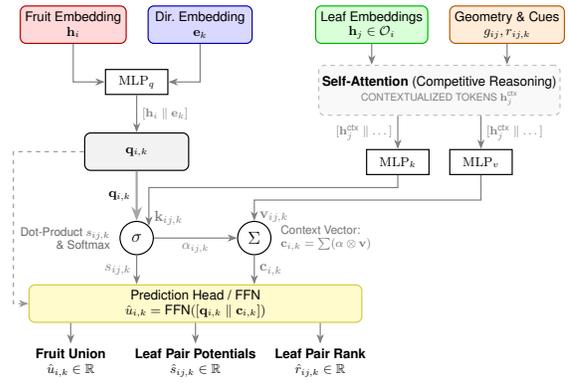
\begin{figure}[t]
\centering
\begin{tikzpicture}[
    >=Stealth,
    node distance=1cm and 0.5cm,
    box/.style={draw, rounded corners, align=center, minimum height=0.9cm, minimum width=2.5cm, font=\sffamily\small, thick},
    proj/.style={draw, rectangle, align=center, minimum height=0.6cm, minimum width=1.5cm, font=\sffamily\small, thick, fill=white},
    circle_op/.style={circle, draw, thick, minimum size=0.8cm, font=\sffamily\large, fill=white},
    fruit_col/.style={fill=red!15, draw=red!70!black},
    dir_col/.style={fill=blue!15, draw=blue!70!black},
    leaf_col/.style={fill=green!15, draw=green!70!black},
    geom_col/.style={fill=orange!15, draw=orange!70!black},
    ffn_col/.style={fill=yellow!25, draw=yellow!80!black}
]

\node[box, fruit_col] (fruit) {Fruit Embedding\\$\mathbf{h}_i$};
\node[box, dir_col, right=0.5cm of fruit] (dir) {Dir. Embedding\\$\mathbf{e}_k$};

\node[box, leaf_col, right=1.5cm of dir] (leaf) {Leaf Embeddings\\$\mathbf{h}_j \in \mathcal{O}_i$};
\node[box, geom_col, right=0.5cm of leaf] (geom) {Geometry \& Cues\\$g_{ij}, r_{ij,k}$};

\coordinate (q_mid) at ($(fruit.south)!0.5!(dir.south)$);
\node[proj, below=0.6cm of q_mid] (q_proj) {$\mathrm{MLP}_q$};

\draw[->, gray, thick] (fruit.south) |- (q_proj.west);
\draw[->, gray, thick] (dir.south) |- (q_proj.east);

\node[box, fill=gray!10, draw=black, below=0.9cm of q_proj] (q_node) {$\mathbf{q}_{i,k}$};
\draw[->, gray, thick] (q_proj.south) -- (q_node.north) node[midway, right, font=\sffamily\footnotesize] {$[\mathbf{h}_i \parallel \mathbf{e}_k]$};

\coordinate (leaf_mid) at ($(leaf.south)!0.5!(geom.south)$);
\node[box, fill=gray!5, draw=gray!50, dashed, below=0.5cm of leaf_mid, minimum width=5.8cm, minimum height=1.2cm] (self_attn_box) {};
\node[align=center, font=\sffamily\small] at (self_attn_box) {\textbf{Self-Attention} (Competitive Reasoning)\\ \color{gray}\scriptsize CONTEXTUALIZED TOKENS $\mathbf{h}^{\text{ctx}}_{j}$};

\draw[->, gray, thick] (leaf.south) -- (leaf.south |- self_attn_box.north);
\draw[->, gray, thick] (geom.south) -- (geom.south |- self_attn_box.north);

\coordinate (K_out) at ($(self_attn_box.south) - (1cm, 0)$);
\coordinate (V_out) at ($(self_attn_box.south) + (1cm, 0)$);

\node[proj, below=0.8cm of K_out] (k_proj) {$\mathrm{MLP}_k$};
\node[proj, below=0.8cm of V_out] (v_proj) {$\mathrm{MLP}_v$};

\draw[->, gray, thick] (K_out) -- (k_proj.north) node[midway, left, font=\sffamily\footnotesize] {$[\mathbf{h}^{\text{ctx}}_{j} \parallel \dots]$};
\draw[->, gray, thick] (V_out) -- (v_proj.north) node[midway, right, font=\sffamily\footnotesize] {$[\mathbf{h}^{\text{ctx}}_{j} \parallel \dots]$};

\node[circle_op, below=1.2cm of q_node] (sigma) {$\sigma$};
\node[left=0.1cm of sigma, font=\sffamily\footnotesize, text=gray!80!black, align=right] {Dot-Product $s_{ij,k}$\\\& Softmax};

\node[circle_op, right=2cm of sigma] (Sigma) {$\Sigma$};
\node[right=0.1cm of Sigma, font=\sffamily\footnotesize, align=left, text=gray!90!black] {Context Vector:\\$\mathbf{c}_{i,k} = \sum(\alpha \otimes \mathbf{v})$};

\draw[->, gray!70!white, line width=1.5pt] (q_node.south) -- (sigma.90) node[pos=0.5, left, text=black, font=\sffamily\small] {$\mathbf{q}_{i,k}$};

\draw[->, gray, thick] (k_proj.south) -- ++(0,-0.3cm) -| (sigma.45) node[pos=0.9, right] {$\mathbf{k}_{ij,k}$};
\draw[->, gray, thick] (v_proj.south) -- ++(0,-0.6cm) -| (Sigma.90) node[pos=0.9, right] {$\mathbf{v}_{ij,k}$};

\draw[->, gray, thick] (sigma.east) -- node[below, font=\sffamily\small] {$\alpha_{ij,k}$} (Sigma.west);

\node[box, ffn_col, minimum width=8cm, below=0.7cm of $(sigma.south)!0.5!(Sigma.south)$] (ffn) {Prediction Head / FFN\\$\hat{u}_{i,k} = \text{FFN}([\mathbf{q}_{i,k} \parallel \mathbf{c}_{i,k}])$};

\draw[->, gray!80!white, thick, dashed] (q_node.west) -- ++(-1.7cm,0) |- (ffn.west);

\draw[->, gray, thick] (sigma.south) -- (sigma.south |- ffn.north) node[pos=0.5, left] {$s_{ij,k}$};
\draw[->, gray, thick] (Sigma.south) -- (Sigma.south |- ffn.north) node[pos=0.5, right] {$\mathbf{c}_{i,k}$};

\node[align=center, below=0.5cm of ffn, xshift=-3cm, font=\sffamily\small] (out_union) {\textbf{Fruit Union}\\$\hat{u}_{i,k} \in \mathbb{R}$};
\node[align=center, below=0.5cm of ffn, xshift=0cm, font=\sffamily\small] (out_pair) {\textbf{Leaf Pair Potentials}\\$\hat{s}_{ij,k} \in \mathbb{R}$};
\node[align=center, below=0.5cm of ffn, xshift=3cm, font=\sffamily\small] (out_rank) {\textbf{Leaf Pair Rank}\\$\hat{r}_{ij,k} \in \mathbb{R}$};

\draw[->, gray, thick] ($(ffn.south) + (-3cm,0)$) -- (out_union.north);
\draw[->, gray, thick] (ffn.south) -- (out_pair.north);
\draw[->, gray, thick] ($(ffn.south) + (3cm,0)$) -- (out_rank.north);
\path[use as bounding box]
  (current bounding box.south west)
  rectangle ([yshift=10pt]current bounding box.north east);
\end{tikzpicture}
\caption{\textbf{Cross-attention architecture for occlusion prediction.}
\scriptsize{A direction-specific fruit query~$\mathbf{q}_{i,k}$ computes pairwise occlusion potentials~$\hat{s}_{ij,k}$ against self-attended, contextualized leaf features.
These potentials act as attention weights to aggregate the leaf features into a single context vector~$\mathbf{c}_{i,k}$, which is then concatenated with the initial query to predict the visibility reduction~$\hat{u}_{i,k}$.}}
\label{fig:occl_attn}
\vspace{-0.3cm}
\end{figure}

Occlusion is inherently directional, but the structural competition among foliage dictates visibility across all viewpoints.
We discretize the visibility space into $K=18$ canonical approach directions $\mathbf{d}_k \in \mathbb{R}^3$, expressed in a predicted fruit-local coordinate frame.
As shown in Fig.~\ref{fig:occl_attn}, we employ a dual-stream cross-attention scorer to efficiently evaluate them against a candidate set of occluding leaves $\mathcal{O}_{i}$.

For each target fruit $i$, we identify $\mathcal{O}_{i}$ using a predefined spatial radius search.
To capture competition and redundancy among nearby foliage, we process the candidate set of leaves using a self-attention encoder once per fruit.
The input tokens concatenate the leaf embedding and direction-independent pair geometry~\mbox{$g_{ij} \in \mathbb{R}^{11}$} (comprising normalized relative displacement, Euclidean distance, and relative bounding box extents and size ratios), producing shared, contextualized leaf tokens $\mathbf{h}^{\text{ctx}}_{j}$ for all $K$ directions.

To evaluate a specific approach direction $k$, we represent the direction with a learned embedding $\mathbf{e}_k$ and form a direction-conditioned fruit query.
Corresponding keys and values are generated by augmenting the contextualized leaf tokens with the pair geometry $g_{ij}$ and direction-specific geometric cues $r_{ij,k} \in \mathbb{R}^2$ (depth and lateral offset along~$\mathbf{d}_k$):
\begin{align}
\mathbf{q}_{i,k} &= \mathrm{MLP}_q([\mathbf{h}_i \parallel \mathbf{e}_k]) \\
\mathbf{k}_{ij,k} &= \mathrm{MLP}_k([\mathbf{h}^{\text{ctx}}_{j} \parallel g_{ij} \parallel r_{ij,k}]) \\
\mathbf{v}_{ij,k} &= \mathrm{MLP}_v([\mathbf{h}^{\text{ctx}}_{j} \parallel g_{ij} \parallel r_{ij,k}])
\end{align}
The pairwise occlusion potential is computed via a scaled dot-product:
\begin{equation}
\vspace{-0.1cm}
\hat{s}_{ij,k} = \frac{\langle \mathbf{q}_{i,k}, \mathbf{k}_{ij,k} \rangle}{\sqrt{D_A}}
\end{equation}
where $D_A$ is the projection dimension.
Leaf evidence is then aggregated via attention pooling, weighted by the temperature-scaled softmax of the potentials:
\begin{equation}
\alpha_{ij,k} = \mathrm{softmax}_{j}\!\left(\frac{\hat{s}_{ij,k}}{T}\right), \quad
\mathbf{c}_{i,k} = \sum_{j \in \mathcal{O}_i} \alpha_{ij,k} \mathbf{v}_{ij,k}
\end{equation}
This context $\mathbf{c}_{i,k}$ is concatenated with the initial query features to predict the union visibility reduction $\hat{u}_{i,k}$, leaf pair potential $\hat{s}_{ij,k}$, and leaf pair rank $\hat{r}_{ij,k}$.
\subsection{Training Objectives}
The network is trained end-to-end using a multi-task objective that balances the heterogeneous supervision signals.

\subsubsection{Node Classification Loss}
We apply standard cross-entropy loss to both the auxiliary local head and the final relational node classifier.
\begin{equation}
\vspace{-0.2cm}
\mathcal{L}_{\mathrm{node}} = -\frac{1}{|\mathcal{V}|}\sum_{i\in\mathcal{V}}\sum_{c=1}^{C} y_{i,c}\log \hat{p}_{i,c}
\end{equation}

\subsubsection{Edge Existence and Semantics Loss}
Because the candidate graph is over-complete, true structural edges are highly sparse.
We address this extreme class imbalance using a weighted binary cross-entropy loss for edge existence:
\begin{equation}
\resizebox{0.89\columnwidth}{!}{$
\mathcal{L}_{\mathrm{exist}} = -\frac{1}{|\mathcal{E}|}\sum_{(u,v)\in\mathcal{E}} \Big( \beta\, y_{u,v}\log \hat{e}_{u,v} + (1-y_{u,v})\log(1-\hat{e}_{u,v}) \Big)
$}
\end{equation}
where $\beta > 1$ upweights the minority positive edges.
For relation type classification, we apply standard cross-entropy restricted strictly to the set of true positive edges $\mathcal{E}^{+}$.
\begin{equation}
\mathcal{L}_{\mathrm{rel}} = -\frac{1}{|\mathcal{E}^{+}|}\sum_{(u,v)\in\mathcal{E}^{+}} \sum_{k=1}^{K} y_{u,v}^{(k)}\log \hat{r}_{u,v}^{(k)}
\end{equation}

\subsubsection{Geometric Regression Loss}
Regressing complete shape parameters from partial point clouds is inherently noisy.
We therefore use a Smooth L1 loss to reduce sensitivity to geometric outliers when predicting centroid offsets and bounding box extents:
\begin{equation}
\scalebox{0.78}{$
\displaystyle
\mathcal{L}_{\mathrm{geom}}
= \frac{1}{|\mathcal{V}|}\sum_{i\in\mathcal{V}}
\Big(
\mathrm{SmoothL1}\!\big(\Delta \hat{\mathbf{c}}_i - \Delta \mathbf{c}_i\big)
+
\mathrm{SmoothL1}\!\big(\hat{\mathbf{s}}_i - \mathbf{s}_i\big)
\Big)
$}
\end{equation}

\subsubsection{Direction-Conditioned Occlusion Losses}
\label{sec:occl_losses}
Training the occlusion module requires balancing global visibility estimation with fine-grained, competitive occluder identification.
To this end, the network is supervised through four complementary objectives: union visibility, pairwise potentials, listwise leaf ranking, and a leaf-to-union consistency regularizer.

\paragraph{Union Visibility Loss}
First, to evaluate global visibility, the union head predicts the visibility reduction ~$\hat{u}_{i,k}$ for each supervised fruit $i \in \mathcal{F}$ and direction $k$, supervised against a soft target $o_{i,k}\in[0,1]$:
\begin{equation}
\resizebox{0.89\columnwidth}{!}{$
\mathcal{L}_{\mathrm{union}}
=
-\frac{1}{|\mathcal{F}|\,K}
\sum_{i,k}
\Big(
o_{i,k}\log \hat{u}_{i,k}
+
(1-o_{i,k})\log\big(1-\hat{u}_{i,k}\big)
\Big)
$}
\end{equation}

\paragraph{Pairwise Potential Loss}
While the union loss captures global visibility, occlusion-aware manipulation requires localizing the specific obstructing organs.
For each candidate leaf $j$, the potential head predicts $\hat{s}_{ij,k}$ supervised against a continuous soft target $p_{ij,k}\in[0,1]$.
To prevent the network from fitting to label noise caused by negligible geometric overlaps, we apply an $\epsilon$-gate using an indicator function:
\begin{equation}
\tilde{p}_{ij,k} = p_{ij,k} \cdot \mathbb{I}(p_{ij,k} \ge \epsilon_{\mathrm{pot}})
\end{equation}
The loss is evaluated only on a selected subset $\mathcal{S}_{\mathrm{pot}}$ of leaves on informative directions per fruit:
\begin{equation}
\resizebox{0.89\columnwidth}{!}{$
\small
\mathcal{L}_{\mathrm{pot}}
=
-\frac{1}{|\mathcal{S}_{\mathrm{pot}}|}
\sum_{i,k,j}
\Big(
\tilde{p}_{i,j,k}\log \hat{s}_{i,j,k}
+
(1-\tilde{p}_{i,j,k})\log\big(1-\hat{s}_{i,j,k}\big)
\Big)
$}
\end{equation}

\paragraph{Listwise Ranking Loss}
Because multiple leaves can jointly occlude a single fruit, treating their potentials independently is insufficient for prioritizing which leaf to push aside first.
To explicitly optimize the retrieval of dominant occluders, we train separate ranking predictions $\hat{r}_{ij,k}$ using a competitive listwise softmax objective over the candidate leaf set.
We define a graded relevance mass $m_{ij,k}$ (set to $0$ whenever $p_{ij,k}<\epsilon_{\mathrm{pot}}$) from the multi-layer depth z-buffer, which yields the target probability distribution~$t_{ij,k}$:
\begin{equation}
\resizebox{0.89\columnwidth}{!}{$
m_{ij,k} = \left(\sum_{n=1}^{3} \mathrm{zbuf}_{n}(i,j,k)\right)^{\gamma}, \quad
t_{ij,k} = \frac{m_{ij,k}}{\sum_{j'} m_{ij',k} + \varepsilon}
$}
\end{equation}
We compute the ranking loss only on fruit--direction queries that are both highly occluded and have nonzero graded relevance mass.
Formally, we define the supervised set
\begin{equation}
\vspace{-0.2cm}
\mathcal{S}_{\mathrm{rank}} = \left\{(i,k) \mid o_{i,k}\ge\tau_{\mathrm{union}},\ \sum_{j} m_{ij,k}>0\right\},
\end{equation}
with $\tau_{\mathrm{union}} = 0.5$.
The listwise objective is then
\begin{equation}
\vspace{-0.1cm}
\mathcal{L}_{\mathrm{rank}}
=
-\frac{1}{|\mathcal{S}_{\mathrm{rank}}|}
\sum_{(i,k)\in\mathcal{S}_{\mathrm{rank}}}
\sum_j
t_{i,j,k}\,
\log \hat{r}_{i,j,k}.
\end{equation}
This objective directly aligns the network with information retrieval metrics, encouraging it to allocate probability mass to the most severe physical obstructions.

\paragraph{Leaf-to-Union Consistency Loss}
To encourage coherence between union visibility and local per-leaf predictions, we introduce a consistency regularizer.
To prevent spurious accumulation from negligible overlaps, we gate this probability using the same threshold $\epsilon_{\mathrm{pot}}$ applied to the ground-truth potentials:
\begin{equation}
\tilde{s}_{i,j,k}
=
\hat{s}_{i,j,k}\,\mathbb{I}\!\left(p_{i,j,k}\ge \epsilon_{\mathrm{pot}}\right)
\end{equation}
We aggregate over the candidate leaf set using a noisy-OR model, which naturally captures the combined effect of independent, overlapping occluders:
\begin{equation}
\hat{u}^{\mathrm{cons}}_{i,k}
=
1
-
\prod_{j}
\left(1-\tilde{s}_{i,j,k}\right)
\vspace{-0.1cm}
\end{equation}
This regularizer is applied only on highly occluded directions that contain at least one gated occluder:
\begin{equation}
\resizebox{0.89\columnwidth}{!}{$
\mathcal{S}_{\mathrm{cons}}
=
\left\{(i,k)\ \middle|\ o_{i,k}\ge\tau_{\mathrm{union}},\ \sum_{j} \mathbb{I}\!\left(p_{i,j,k}\ge \epsilon_{\mathrm{pot}}\right) > 0\right\}
$}
\end{equation}
We regress the aggregated estimate toward the continuous union target using mean squared error:
\begin{equation}
\vspace{-0.1cm}
\resizebox{0.89\columnwidth}{!}{$
\mathcal{L}_{\mathrm{cons}}
=
\frac{1}{|\mathcal{S}_{\mathrm{cons}}|}
\sum_{(i,k)\in\mathcal{S}_{\mathrm{cons}}}
\left\|
\hat{u}^{\mathrm{cons}}_{i,k}
-
o_{i,k}
\right\|_2^2
$}
\end{equation}
This term encourages predicted global visibility reduction to be supported by identifiable candidate occluders.
Finally, the total occlusion objective combines these global, local, and competitive signals using the respecting weighting factors $\lambda$'s:
\begin{equation}
\resizebox{0.89\columnwidth}{!}{$
\mathcal{L}_{\mathrm{occl}}
= \lambda_{\mathrm{union}}\mathcal{L}_{\mathrm{union}}
+ \lambda_{\mathrm{pot}}\mathcal{L}_{\mathrm{pot}}
+ \lambda_{\mathrm{rank}}\mathcal{L}_{\mathrm{rank}}
+ \lambda_{\mathrm{cons}}\mathcal{L}_{\mathrm{cons}}
$}
\end{equation}

\section{Experiments}
\label{sec:exp}
We evaluate SG-DOR on a synthetic pepper plant dataset to validate its structured perception and direction-conditioned occlusion reasoning, including its robustness to geometric noise and viewpoint jitter.
Furthermore, we validate our occlusion metrics across different ray-casting methods and provide qualitative real-world validation using a physical mock-up~\cite{isaacjose2025iros}.

\begin{table*}[t]
\scriptsize
\centering
\begin{adjustbox}{max width=\textwidth, margin=0pt 0pt 0pt 3pt}

\begin{tabular}{lllccccccccccc}
\toprule
\textbf{Type}
&\textbf{Model}
&\textbf{Key}
& \shortstack{\textbf{NDCG}\\@3$\uparrow$}
& \shortstack{\textbf{Recall}\\@1$\uparrow$}
& \shortstack{\textbf{Recall}\\@3$\uparrow$}
& \shortstack{\textbf{MAE}\\($\mathbf{U}_{high}$)\,$\downarrow$}
& \shortstack{\textbf{MAE}\\($\mathbf{U}_{low}$)\,$\downarrow$}
& \shortstack{\textbf{MAE}\\($\mathbf{U}_{mid}$)\,$\downarrow$}
& \shortstack{\textbf{Occl Dir}\\F1$\uparrow$}
& \shortstack{\textbf{Edge Exist}\\F1$\uparrow$}
& \shortstack{\textbf{MAE}\\($\Delta\mathbf{c}$)\,$\downarrow$}
& \shortstack{\textbf{MAE}\\($\Delta\mathbf{e}$)\,$\downarrow$}\\
\midrule
\multirow{2}{*}{Baseline}
& SG-DOR-w/o SelfAttn &B0 & 0.567 & 0.277 & 0.556 & 0.240 & 0.114 & 0.123 & 0.448 & \textbf{0.842} & 0.018 & 0.024 \\
& SG-DOR-w/o PairGeom &B1 & 0.639 & 0.321 & 0.624 & 0.181 & 0.130 & \textbf{0.116} & 0.591 & 0.833 & 0.018 & 0.028 \\
\midrule
\multirow{1}{*}{Ours}
&\textbf{SG-DOR} &M0  & \textbf{0.851} & \textbf{0.460} & \textbf{0.818} & \textbf{0.109} & 0.089 & 0.120 & \textbf{0.734} & 0.832 & 0.015 & 0.020 \\
\midrule
\multirow{2}{*}{Attn Abl}
& SG-DOR-SAOneLayer &A0 & 0.821 & 0.441 & 0.792 & 0.130 & 0.089 & 0.120 & 0.696 & 0.825 & 0.016 & 0.024 \\
& SG-DOR-SAPerDir &A1 & 0.817 & 0.435 & 0.792 & 0.141 & \textbf{0.078} & 0.117 & 0.693 & 0.796 & \textbf{0.014} & \textbf{0.019} \\
\midrule
\multirow{2}{*}{Feat Abl.}
& SG-DOR-GNNOnly &A2 & 0.841 & 0.455 & 0.810 & 0.110 & 0.089 & 0.122 & 0.727 & 0.820 & 0.016 & 0.022 \\
& SG-DOR-EncOnly & A3 & 0.834 & 0.448 & 0.805 & 0.115 & 0.091 & 0.121 & 0.719 & 0.810 & 0.015 & 0.020 \\
\midrule
\multirow{5}{*}{Loss Abl}
& SG-DOR-w/oRankLoss &A4 & 0.787 & 0.416 & 0.765 & 0.116 & 0.110 & 0.130 & 0.698 & 0.842 & 0.016 & 0.022 \\
& SG-DOR-w/oPotLoss &A5 & 0.795 & 0.421 & 0.768 & 0.113 & 0.090 & 0.124 & 0.719 & 0.839 & 0.017 & 0.023 \\
& SG-DOR-w/oConstLoss &A6 & 0.845 & 0.462 & 0.812 & 0.110 & 0.091 & 0.125 & 0.727 & 0.827 & 0.015 & 0.019 \\
& SG-DOR-w/oUnionLoss &A7 & 0.842 & 0.454 & 0.813 & 0.148 & 0.445 & 0.166 & 0.356 & 0.838 & 0.015 & 0.019 \\
& SG-DOR-UnionLossOnly &A8 & 0.494 & 0.245 & 0.497 & 0.180 & 0.162 & 0.123 & 0.546 & 0.809 & 0.020 & 0.026 \\
\bottomrule
\end{tabular}
\end{adjustbox}
\caption{
\textbf{Baseline comparisons and ablation study of the SG-DOR architecture.}
\scriptsize
{Retrieval metrics (NDCG@3, R@1, R@3) evaluate the model's ability to correctly identify and rank dominant occluding leaves.
Union visibility error (MAE) is stratified by ground-truth occlusion severity ($\mathbf{U}_{high}$, $\mathbf{U}_{mid}$, $\mathbf{U}_{low}$). Geometric regression errors for centroids and extents are denoted by $\Delta\mathbf{c}$ and $\Delta\mathbf{e}$.
\textbf{Key findings:} The full proposed model (\textbf{SG-DOR}) achieves the best balance of retrieval accuracy and visibility estimation.
Removing competitive relational reasoning (\textbf{w/oSelfAttn}) or explicit spatial cues (\textbf{w/oPairGeom}) causes substantial drops in leaf ranking (NDCG@3) and severe errors in highly occluded viewpoints ($\mathbf{U}_{high}$).
Furthermore, the loss ablations show that relying solely on global visibility supervision (\textbf{SG-DOR-UnionLossOnly}) leads to a complete collapse in retrieval performance, confirming the proposed listwise ranking and pairwise potential objectives are critical.}
}\label{tab:ablations}
\vspace{-0.3cm}
\end{table*}

\subsection{Implementation Details}
\label{sec:implementation}
We trained the models for 100 epochs using AdamW (weight decay $10^{-4}$) with an initial learning rate of $10^{-3}$ and cosine annealing ($T_{\max}=100$), and with training and validation batch sizes of 8 and 1, respectively.
We used a staged curriculum: local semantic head only for epochs 0-5, geometric regression added at epoch 6, and structural and occlusion heads at epoch 11, after which the full multi-task objective was trained jointly.

As described in (\secref{sec:dataset_generation}), we generated 800 synthetic scenes with an 80/20 scene-level train/validation split, and extracted point cloud from meshes at 4mm voxel resolution.
Unless stated otherwise, we applied on-the-fly augmentation: Gaussian jitter ($\sigma=2$ mm) and 20\% point dropout (with replacement) after instance sampling.
We used $N_p=128$ points as input to the PointNet++ encoder.
The relational backbone is a 4-layer residual GINE with hidden width 128 and dropout 0.1.

For occlusion reasoning, each fruit is evaluated over \mbox{$K=18$ directions} in its local frame, with candidate leaves selected within a $0.2\,\mathrm{m}$ radius.
The occlusion module uses a 16-dimensional direction embedding and a 2-layer leaf-set self-attention encoder (width 128, 4 heads), followed by attention pooling ($T=1.0$).
\subsection{Metrics}
\label{sec:metrics}
\noindent\textbf{Leaf Retrieval:}
We evaluate dominant-occluder identification over the candidate leaf set per fruit-direction $(i,k)$ query using NDCG@3, Recall@1 (R@1), and Recall@3 (R@3).
Specifically, R@1 and R@3 measure the frequency with which the primary occluding leaf is correctly retrieved within the top one or three predictions, respectively, while NDCG@3 evaluates the ranking quality of the top three candidates.
NDCG uses linear gain with a $\log_2$ rank discount, where the ground-truth graded relevance of each leaf is directly quantified by its actual occlusion severity, defined as its accumulated z-buffer mass given by~\eqref{eq:zbuf}.

\noindent\textbf{Union Visibility Reduction:}
For each fruit-direction $(i,k)$, we regress the continuous union visibility reduction score in $[0,1]$ and report mean absolute error~(MAE) in three bins based on ground-truth union: $\mathbf{U}_{low}$ for $<0.25$, $\mathbf{U}_{high}$ for $>0.5$, and $\mathbf{U}_{mid}$ for values in between.
To evaluate the binary classification of dominant directional occlusions, we report the maximum global \mbox{$F_1$-score} (\textbf{Occl Dir F1}).
The ground-truth binary label for a specific fruit-direction pair is defined as $y_{i,k} = \mathbb{1}[U_{i,k} \ge \tau]$, where $U_{i,k}$ is the union occlusion. Crucially, this metric is computed globally and jointly across all fruit-direction pairs.

\noindent\textbf{Scene Graph:}
Edge Exist F1 denotes the best F1 score for binary edge existence over candidate edges.
Geometric accuracy is measured as MAE of centroid offsets $\Delta\mathbf{c}$ and extent residuals $\Delta\mathbf{e}$ per node.
Note that semantic node prediction and edge relation type prediction for positive edges have $F1>0.99$ for all models and hence we do not explicitly report them.

\subsection{Baselines and Ablation Studies}
\label{sec:ablations}
Because direction-conditioned listwise occlusion ranking is a novel task formulation, direct off-the-shelf baselines are not available for comparison.
Instead, we construct representative baselines by adapting the SG-DOR architecture to mimic standard 3D perception paradigms, and subsequently ablate individual feature and loss components to validate our design (Table~\ref{tab:ablations}).

\paragraph{Baseline Comparison}
We formulate two primary baselines to represent conventional approaches to set processing and graph reasoning.
First, the \textbf{SG-DOR-w/oSelfAttn} baseline removes the transformer encoder from the leaf set(see~\figref{fig:occl_attn}). In this modified architecture, leaf tokens are processed independently via standard MLPs before being aggregated, mirroring conventional PointNet-style set processing that lacks competitive relational reasoning.
Second, the \textbf{SG-DOR-w/oPairGeom} baseline removes explicit fruit-local relative geometry ($g_{ij}$) from the input tokens, forcing the network to infer spatial relationships entirely from the latent GNN node embeddings. This represents standard semantic scene graph networks that do not inject explicit, task-specific relative geometry.

Both conventional baselines fail to perform the occlusion retrieval task effectively.
Relying on independent leaf processing (\textbf{SG-DOR-w/oSelfAttn}) causes a significant drop in retrieval performance, lowering NDCG@3 from 0.851 to 0.567 and Recall@1 from 0.460 to 0.277.
Similarly, omitting explicit pairwise geometry (\textbf{SG-DOR-w/oPairGeom}) degrades NDCG@3 to 0.639 and significantly increases the mean absolute error for highly occluded viewpoints ($\mathbf{U}_{high}$ MAE: 0.181 vs. 0.109).
These results confirm that standard independent point-set processing and purely latent spatial reasoning are insufficient. Explicit geometric cues and relational contextualization among leaves are necessary to accurately resolve physical occlusion redundancies.

\paragraph{Attention Mechanisms and Feature Representations}
We compare the proposed direction-independent per-fruit leaf-set encoding against a per-fruit per-direction variant (\textbf{SG-DOR-SAPerDir}), where self-attention is computed separately for each of the $K=18$ query directions.
The per-direction approach yields lower retrieval performance (NDCG@3: 0.817 vs. 0.851), suggesting that computing a shared leaf context provides a regularizing effect that limits overfitting to specific directional queries.
Furthermore, a single layer of self-attention on leaves (\textbf{SG-DOR-SAOneLayer}) proved to be insufficient with NDCG@3 and Occl Dir F1 dropping to 0.821 and 0.696.
Regarding input representations, restricting the node features to solely GNN-refined embeddings (\textbf{SG-DOR-GNNOnly}) or solely raw PointNet++ features (\textbf{SG-DOR-EncOnly}) results in a measurable decrease in overall performance.
This implies that combining intrinsic organ geometry with broader neighborhood topology provides the most effective context for the cross-attention scorer.

\paragraph{Evaluation of the Multi-Task Objective}
Ablating the loss components isolates the contribution of each supervision signal.
Relying exclusively on the global union visibility loss (\textbf{SG-DOR-UnionLossOnly}) yields the lowest overall ranking performance (NDCG@3: 0.494) as well as a very low Occl Dir F1 of 0.546, demonstrating that global occlusion estimation benefits from the constraints provided by fine-grained, per-leaf supervision.
Conversely, omitting the union loss (\textbf{SG-DOR-w/oUnionLoss}) preserves the relative ranking ability but severely degrades absolute directional visibility estimation ($\mathbf{U}_{low}$ MAE increases to 0.445; Occl Dir F1 drops to 0.356).
Excluding the listwise ranking objective (\textbf{SG-DOR-w/oRankLoss}) and the pairwise potential loss (\textbf{SG-DOR-w/oPotLoss}) both reduce retrieval performance, with NDCG@3 dropping to 0.787 and 0.795 respectively, indicating that while both contribute to accurate occluder ordering, the listwise ranking objective plays the more critical role in prioritizing dominant occluders.

\begin{table}[tb]
\centering
\begin{adjustbox}{margin=0pt 0pt 0pt 3pt}
\begin{tabular}{cccccc}
\toprule
 \shortstack{\textbf{Model}\\\textbf{Key}} & \shortstack{\textbf{NDCG}\\@3$\uparrow$} & \shortstack{\textbf{Occl Dir}\\F1$\uparrow$} & \shortstack{\textbf{Edge Exist}\\F1$\uparrow$} & \shortstack{\textbf{MAE}\\($\Delta\mathbf{c}$)$\downarrow$} & \shortstack{\textbf{MAE}\\($\Delta\mathbf{e}$)$\downarrow$} \\
\midrule
M0 & 0.728 & 0.634 & 0.832 & 0.019 & 0.034 \\
A1 & 0.713 & 0.526 & 0.865 & 0.022 & 0.041 \\
A2 & 0.760 & 0.602 & 0.627 & 0.018 & 0.038 \\
A3 & 0.740 & 0.616 & 0.860 & 0.018 & 0.039 \\
A6& 0.585 & 0.524 & 0.739 & 0.029 & 0.045 \\
\bottomrule
\end{tabular}
\end{adjustbox}
 \caption{\textbf{Stress tests with XYZ jitter of $4$\,mm during training.}
 \scriptsize{Refer to~\tabref{tab:ablations} for key to model mapping. Notably, under severe geometric noise, local point features help preserve structural topology (A2 vs.\ A3 on \textbf{Edge Exist F1}), while the consistency loss becomes important for stabilizing occlusion retrieval (evidenced by the drop in \textbf{NDCG@3} and \textbf{Occl Dir F1} for A6 compared to M0).}}
\label{tab:stress}
\vspace{-0.5cm}
\end{table}
\subsection{Robustness Under Severe Geometric Noise}
To evaluate representation learning under severe spatial uncertainty, models were trained with $4$\,mm XYZ coordinate jitter and evaluated on clean data~(\tabref{tab:stress}).
Relying solely on GNN-refined features (\textbf{A2}) severely degrades test-time edge prediction (Edge Exist F1: 0.627), as persistent graph corruption during training prevents the network from learning valid structural priors. Conversely, the raw PointNet++ encoder~(\textbf{A3}) isolates jitter-invariant shape features, maintaining high structural accuracy (Edge Exist F1: 0.860).
The full model (\textbf{M0}) effectively balances these representations, preserving topology while achieving better directional occlusion accuracy (Occl Dir F1: 0.634 vs.\ 0.616) than~\textbf{A3}.
Furthermore, computing self-attention independently per direction~(\textbf{A1}) fails to learn generalized visibility estimates (Occl Dir F1: 0.526).
Finally, while leaf-to-union consistency (\textbf{A6}) appeared redundant in \tabref{tab:ablations}, its absence under noisy training causes a retrieval collapse (NDCG@3: 0.585), demonstrating its critical role in enforcing valid physical occlusion logic when geometric training signals are highly ambiguous.

\subsection{Comparison under Viewpoint and Projection Distortion}
\begin{table}[b]
\centering
\begin{adjustbox}{margin=0pt 0pt 0pt 3pt}
\begin{tabular}{lcccc}
\toprule
\textbf{GT Variant} & \shortstack{\textbf{Occl Dir} F1$\uparrow$} &  \shortstack{\textbf{NDCG}@3$\uparrow$}  &   \shortstack{\textbf{MAE}($\mathbf{U}_{high}$)\,$\downarrow$}\\
\midrule
\texttt{zbuf} & 0.775 & 0.898 & 0.100 \\
\midrule
\texttt{ortho\_j0} & 0.739  & 0.890 & 0.111 \\
\texttt{ortho\_j1} & 0.706 & 0.876  & 0.122 \\
\midrule
\texttt{persp\_j0} & 0.689 & 0.855 & 0.262 \\
\texttt{persp\_j1} & 0.652 & 0.834 & 0.278 \\
\bottomrule
\end{tabular}
\end{adjustbox}
\caption{
\textbf{Impact of ground truth modality and viewpoint misalignment on occlusion reasoning.}
\scriptsize{We evaluate the model (trained on Z-buffer) against ray-cast simulations using orthographic and perspective projections.
Row 1 (\texttt{zbuf}) uses the original Z-buffer ground truth as the validation metric.
Subsequent rows evaluate generalization to orthographic and perspective camera-based ray-casting at canonical angles ($j_0$) and with a $5^\circ$ angular viewpoint jitter ($j_1$).}
}
\label{tab:camera_tests}
\vspace{-0.5cm}
\end{table}
To validate SG-DOR's geometric reasoning, we evaluate the model, trained exclusively on z-buffer data (\secref{sec:dataset_generation}), against ground truth metrics obtained from ray-casting simulations featuring orthographic and perspective projection shifts and angular jitter (\tabref{tab:camera_tests}).
The \texttt{zbuf} row serves as the training-modality validation metric.
Under un-jittered orthographic ray-casting (\texttt{ortho\_j0}), performance tightly tracks the baseline (NDCG@3: 0.890 vs. 0.898), indicating the model learns genuine occlusion structures rather than overfitting to z-buffer artifacts.
Even under challenging perspective distortion (\texttt{persp\_j0}), it maintains ranking integrity (NDCG@3 = 0.855) despite expected spatial error increases from non-linear scaling of perspective projection (\textbf{MAE}($\mathbf{U}_{high}$) of 0.262 versus 0.100).
Finally, a $5^\circ$ viewpoint perturbation (\texttt{j1}) causes only graceful degradation, confirming the network captures robust volumetric relationships that persist beyond canonical training views. These experiments additionally provide geometry-based comparisons using alternative viewpoint and projection formulations.

\subsection{Qualitative Real-World Evaluation}
\begin{figure}[t]
\centering
\resizebox{0.85\columnwidth}{!}{\input{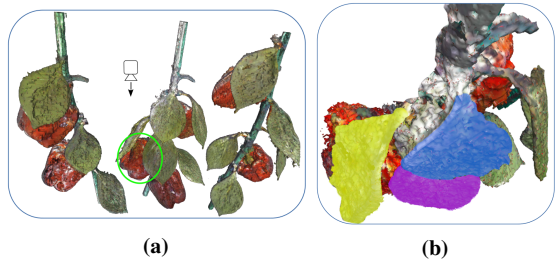}}
\caption{\textbf{Qualitative Real-World Evaluation}
\scriptsize{Fig. 5a shows the mock-up of sweet pepper plants, with the fruit of interest encircled in green and the viewing direction indicated from above. Fig. 5b shows the view along the approach direction, three predicted occluding leaves highlighted in yellow, purple, and blue, corresponding to ranks \#1--\,\#3 (most to least relevant occluders).}}
\label{fig:qual_real}
\vspace{-0.55cm}
\end{figure}
To further validate our approach in physical scenarios, we applied SG-DOR to a sweet pepper plant mock-up~\cite{lenz2024hortibot} and a reconstructed 3D model~\cite{isaacjose2025iros}, which we additionally instance-segmented into individual organs.
Despite dealing with incomplete real-world 3D reconstructions, sensor noise, and dense foliage, the model successfully identified and ranked occluding leaves without any real-world fine-tuning.
As illustrated in \figref{fig:qual_real}a, the target fruit (encircled in green) presents a  challenging scenario surrounded by multiple occluder candidates.
Yet, for a top-down approach direction, SG-DOR correctly isolates and ranks the three primary occluding leaves (highlighted in yellow, purple, and blue in \figref{fig:qual_real}b).
These qualitative results suggest that our method can zero-shot learn physical representations suitable for real-world robotic manipulation.

\section{Conclusion}
We introduced SG-DOR, a scene-graph framework that jointly learns organ semantics, attachment structure, geometry, and direction-conditioned occlusion reasoning for sweet pepper plants.
A key contribution is the occlusion ranking task: given a fruit and approach direction, the model retrieves and orders candidate leaves by graded relevance in a fruit-local frame.
We implemented SG-DOR via per-fruit leaf-set self-attention and a direction-conditioned cross-attention scorer with listwise supervision.
Experiments demonstrate that SG-DOR reliably infers direction-conditioned occlusion relations and preserves structural attachments, with occlusion results independently verified against a ray-casting simulation.
These predictions can serve as actionable cues for targeted leaf removal in autonomous horticulture systems.

\section{Acknowledgements}
Generative AI, namely GPT-5.6, was used for text proofreading, and Gemini 3 Pro for the generation of \figref{fig:full_pipeline}.
\bibliographystyle{IEEEtran}
\balance
\bibliography{references}

\end{document}